\documentclass[10pt,twocolumn,letterpaper]{article}

\usepackage{cvpr}              

\usepackage{graphicx}
\usepackage{amsmath}
\usepackage{amssymb}
\usepackage[symbol]{footmisc}

\usepackage{booktabs}
\usepackage{comment}
\usepackage{multirow}
\usepackage[accsupp]{axessibility}

\usepackage[pagebackref,breaklinks,colorlinks]{hyperref}

\usepackage[capitalize]{cleveref}
\crefname{section}{Sec.}{Secs.}
\Crefname{section}{Section}{Sections}
\Crefname{table}{Table}{Tables}
\crefname{table}{Tab.}{Tabs.}

\def\cvprPaperID{11} 
\def\confName{CVPR}
\def\confYear{2022}

\begin{document}

\title{Detecting, Tracking and Counting Motorcycle Rider Traffic Violations on Unconstrained Roads}
\author{Aman Goyal\footnotemark[1] \hspace{25mm} Dev Agarwal\footnotemark[1]
\\Anbumani Subramanian \hspace{5mm} C.V. Jawahar \hspace{5mm} Ravi Kiran Sarvadevabhatla \hspace{5mm} Rohit Saluja 
\\
\\Centre For Visual Information Technology (CVIT), IIIT Hyderabad, INDIA\\
{\tt\normalsize \url{https://github.com/iHubData-Mobility/public-motorcycle-violations}}
}
\maketitle
\footnotetext[1]{ Authors contributed equally to this research.
\\
{\tt\small \{aman.goyal1099, devagarwal.iitkpg\}@gmail.com},
{\tt\small \{anbumani, jawahar, ravi.kiran\}@iiit.ac.in},
{\tt\small rohit.saluja@reasearch.iiit.ac.in}}
\begin{abstract}
   In many Asian countries with unconstrained road traffic conditions, driving violations such as not wearing helmets and triple-riding are a significant source of fatalities involving motorcycles. Identifying and penalizing such riders is vital in curbing road accidents and improving citizens' safety. With this motivation, we propose an approach for detecting, tracking, and counting motorcycle riding violations in videos taken from a vehicle-mounted dashboard camera. We employ a curriculum learning-based object detector to better tackle challenging scenarios such as occlusions. We introduce a novel trapezium-shaped object boundary representation to increase robustness and tackle the {\it rider-motorcycle} association. We also introduce an amodal regressor that generates bounding boxes for the occluded riders. Experimental results on a large-scale unconstrained driving dataset demonstrate the superiority of our approach compared to existing approaches and other ablative variants.
\end{abstract}
\vspace{-4mm}
\section{Introduction}
\label{sec:intro}
Automated road surveillance has become increasingly crucial as road crashes have become the $8^{th}$ leading cause of death worldwide. A World Health Organization study on road safety~\cite{world2018global,morth2019RA} claims that violations lead to $1.35$ million fatalities and affect $50$ million people yearly. Another recent report by  World Bank~\cite{world2021traffic} mentions that more than $50\%$ of road fatalities involve two-wheeler vehicles, also showing that `no helmet' and triple-riding (more than two riders) violations are common causes. Studies carried out in Asian countries also account for two-wheeler vehicles among the significant share for road fatalities~\cite{morth2019RA,pervez2021identifying,xiong2016risk}.   
Often, static cameras may not be present on the majority of the streets. Installing cameras everywhere may not be an economically viable and sustainable solution. Therefore, we adopt an approach that operates on videos from a dashboard camera mounted within a police vehicle.

\begin{figure}[t]
  \centering
  {\includegraphics[width=\linewidth]{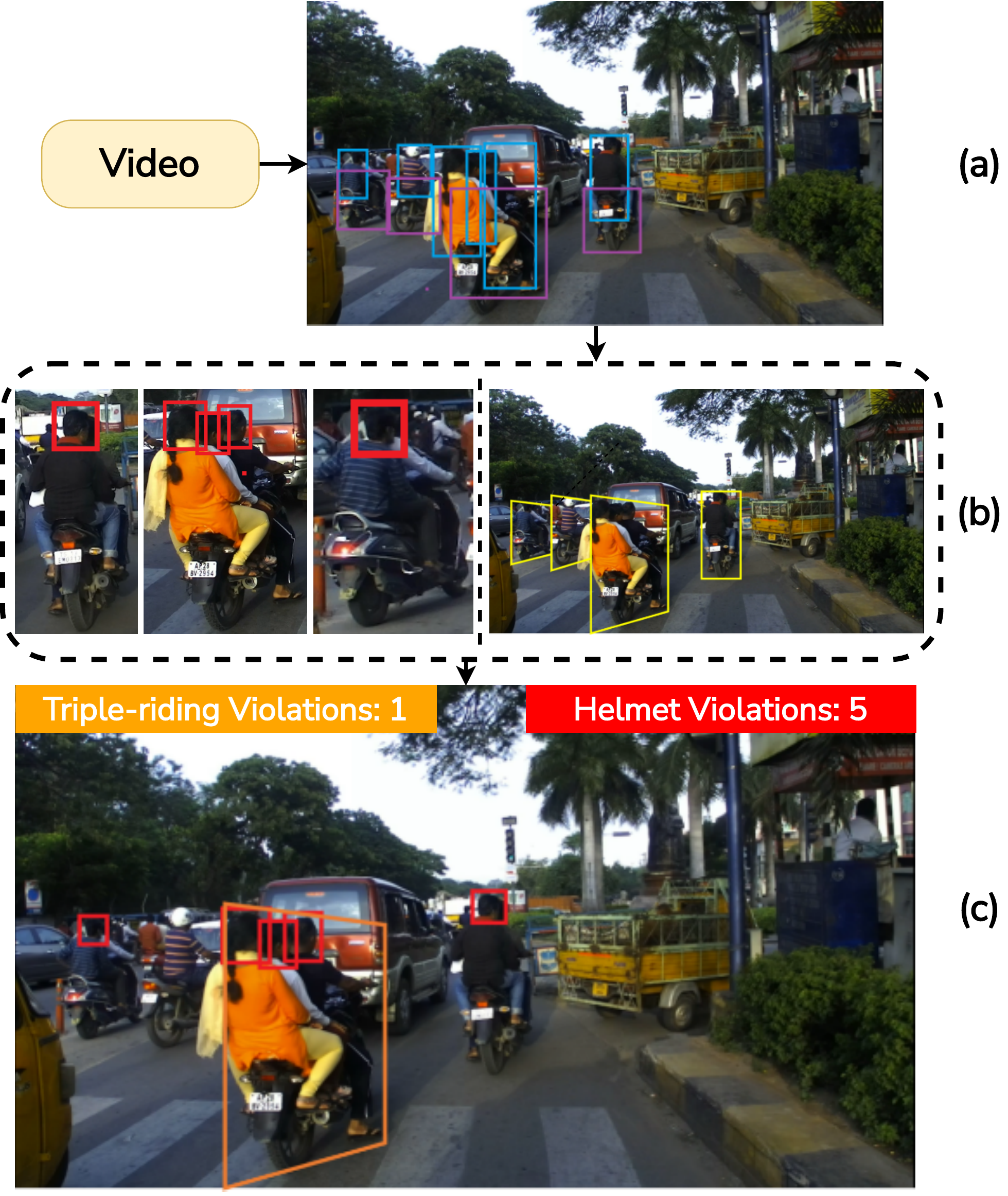}
  }
  \vspace{-5mm}
  \caption{(a) Frame level rider and motorcycle predictions from our curriculum learning-based detector, (b) Outputs from violation detection and counting module, (c) Violation detections and counts overlaid on the frame; triple-riding and helmet violations in orange trapezium (note rider/head occlusions) and red boxes. Refer ~\cref{Figure:2} for remaining color schemes.}
\label{Figure:1}\vspace{-5mm}
\end{figure}

Existing methods for triple-riding violations~\cite{mallela2021detection,saumya2020machine} involve heuristic rules applied to outputs of rider detectors and show qualitative results for a few samples. In addition, these methods are not robust to occlusions and crowded scenes. In contrast, we propose a novel trapezium representation for identifying triple riding violations. The representation largely reduces false positives and is robust as it does not require any heuristic rules. For helmet violations, a majority of the existing works~\cite{Dahiya16,Shine2020,Singh2020,Chairat_2020_WACV,khan2020helmet} perform classification over upper portions of predicted riders. Due to the unavailability of context in the Regions Of Interest (ROIs), such works are not robust in crowded road scenarios. Our proposed technique performs well even in crowded scenarios, as we use the ROIs of motorcycle-driving instances containing sufficient context for {\it helmet/no-helmet} detector to provide accurate results. As an integral part of our pipeline, we use a curriculum learning-based detector to predict motorcycle and rider boxes to overcome inter-class confusion due to overlapping regions among the two classes. Recent methods in curriculum learning mainly focus on object classification~\cite{hacohen2019power,mousavi2021deep,wang2019dynamic} with very few works on detection~\cite{singhorder,wang2018weakly} which involve handling intra-class scale variations or weakly and semi-supervised training.

\begin{figure*}[!ht]
  \centering
  {%
  \includegraphics[width=\linewidth]{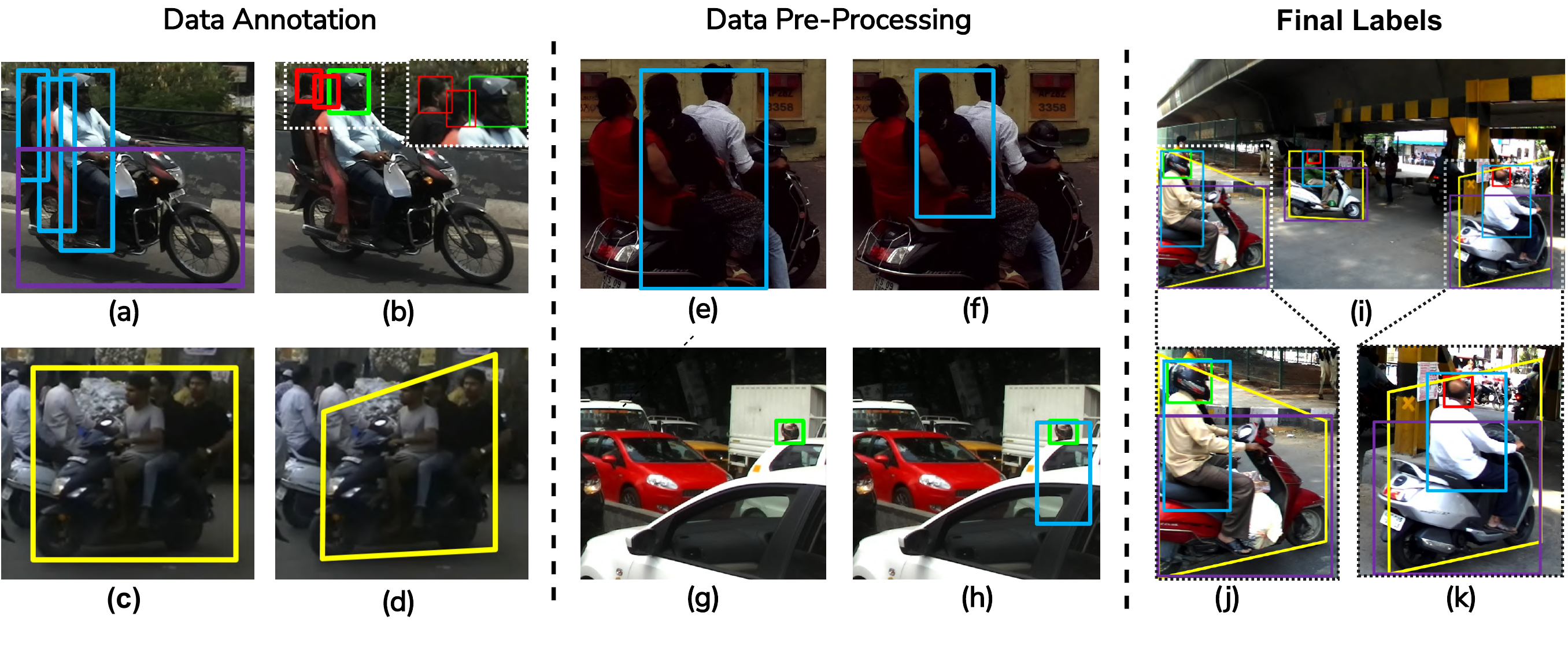}}
  \vspace{-9mm}
  \caption{(a) Existing annotations of rider/motorcycle in IDD dataset, (b) {\it helmet/no-helmet} annotations additionally annotated by us, (c) Conventional {\it rider-motorcycle} instance representation. (d) Novel trapezium box representation for the {\it rider-motorcycle} instance in contrast to (c); reduces false-positives from nearby riders, (e) Rider box annotation present in IDD dataset, (f) Manually corrected rider boxes, (g) Annotated helmet box on an occluded rider, 
(h) Generating boxes for occluded riders with amodal regressor using helmet boxes (refer ~\cref{Section:Dataset}), (i) Sample frame with final labels \& its crops (j), (k). [color scheme: red-helmet violation, green-helmet, orange-triple-riding violation, yellow-driving instance, purple-motorcycle, blue-rider]}
  \label{Figure:2}\vspace{-2mm}
\end{figure*}
To the best of our knowledge, no existing solution jointly tackles helmet and triple-riding violations for crowded scenes and provides robustness in occluded rider scenarios. ~\cref{Figure:1} shows outputs of various components of the proposed architecture wherein we train a curriculum learning-based model to predict motorcycles and riders over the video frame followed by a violation detection and counting module. Our contributions in this paper are:

\begin{enumerate}
    \item A novel dataset pre-processing pipeline involving an amodal regressor to generate boundary representations even for occluded riders.
    \item An innovative trapezium-shaped box regressor for associating a motorcycle with its corresponding riders.
    \item A curriculum learning-based architecture to jointly detect, track and count triple-riding and helmet violations on unconstrained road scenes.
\end{enumerate}
\section{Related Works}
\label{Section:Related Works}
\noindent \textbf{Triple-riding Violations:} There are very few existing methods in the space of triple-riding violations. The method proposed by~\cite{mallela2021detection} detects violation on the lower portion of the motorcycle, is not robust to false positives as no counting algorithm is involved, and all the images used for qualitative analysis have similar camera angles and lighting conditions. The work by~\cite{saumya2020machine} focuses on using an association algorithm based on Euclidean distance after detecting motorcycle and rider boxes to count the number of riders on a motorcycle. It fails in fairly crowded road scenarios and lacks qualitative analysis on a large-scale dataset. However, our proposed approach based on data pre-processing to handle occlusions and trapezium-shaped bounding boxes can robustly detect and track triple-riding violations in each video frame.

\noindent \textbf{Helmet Violations:} Existing works on helmet violations by \cite{Dahiya16,Shine2020,Singh2020,Chairat_2020_WACV,khan2020helmet} classify the ROIs generated from the upper portion of the detected riders. Such approaches are not robust to false negatives and false positives in the form of truncated heads in the cropped ROIs and riders of other motorcycles, respectively. The work by \cite{vishnu2017detection} uses conventional techniques utilizing hand-engineered features such as scale-invariant feature transform, histogram of gradients etc.~\cite{lowe2004distinctive,dalal2005histograms,guo2010completed,cortes1995support} do not perform well on large-scale unconstrained and crowded road scenarios. Works of~\cite{kumar2020automatic} focus on detecting helmet violations directly from complete scene images. Such an approach leads to more false negatives and cannot detect violations for riders farther away from the camera. Apart from this, work by \cite{yogameena2019deep} requires foreground segmentation as a step before detecting {\it helmets/no-helmets}. We obtain robust performance by involving ROIs containing motorcycles and corresponding riders. Though works such as \cite{Chairat_2020_WACV,harish2021evaluate, lin2021positional, jia2021real}, robustly detect helmet violations, they do not possess a triple-riding detection module. On the contrary, our proposed approach can jointly detect, track, and count triple-riding and helmet violations.

\textbf{Curriculum Learning:} The researchers have widely focused on applying curriculum learning-based techniques to object classification~\cite{hacohen2019power,mousavi2021deep,wang2019dynamic}. Existing curriculum-learning-based detection works~\cite{singhorder,wang2018weakly} either concentrate on improving the performance by handling intra-class scale variations or involve weakly and semi-supervised training. Our method utilizes the curriculum learning-based object detector to obtain robust detections of motorcycles and riders and avoid inter-class confusion due to high overlap between the bounding boxes of the two classes. The object detector further helps in robustly predicting triple riding and helmet violations. 

\section{Dataset}~\label{Section:Dataset}
We train various models for identifying motorcycle violations. We use a subset of India Driving Dataset (IDD)~\cite{varma2019idd} extracted by selecting images with motorcycles and riders followed by filtering out small bounding boxes based on an area threshold of $900$ squared pixels, as low-resolution boxes add noisy samples to the training data. We annotate {\it helmet class, no-helmet class}, and {\it trapezium-shaped driving instance class} (see ~\cref{Figure:2} b-d, i-k). We use conventional rectangular boxes for helmet violations and propose trapezium bounding boxes to detect triple-riding violations. We now discuss how we pre-process the dataset.

{\bf Data Pre-Processing:}
We note two crucial issues in the existing IDD dataset annotations, which pose a problem in accurately detecting the riders and subsequent triple-riding and helmet violations. i) Some IDD samples had multiple riders in a single bounding box (due to inclusion of riders' legs), as shown in ~\cref{Figure:2} (e). ii) There were frequent rider occlusions, to the extent that only the head portion is visible, as shown in ~\cref{Figure:2} (g). We overcome such issues by:

{\bf (i)} Processing the Rider Boxes: Rider bounding boxes are manually processed such that each bounding box contains only one rider. ~\cref{Figure:2} (f) shows the output of such processing for a rider in ~\cref{Figure:2} (e).

{\bf (ii)} Generating Boxes for Rider Occlusions with Amodal Regressor: To obtain full bounding boxes for occluded riders, we train a two-layered deep network with fully connected layers, which we refer to as amodal regressor. With {\it helmet/no-helmet} bounding box as input and corresponding {\it rider} bounding box as output, the amodal regressor has $16$ and $64$ nodes in the respective hidden layers.  
We train it on non-occluded riders in our data with {\it relu activation} and {\it learning rate} of $0.001$. We use the trained model to generate boxes for occluded riders in our data. ~\cref{Figure:2} (h) depicts the example of a bounding box generated for an occluded rider.

\begin{figure*}[!ht]
  \centering
  \includegraphics[trim={-0.5cm -0.5cm -0.5cm -0.5cm},clip,width=\linewidth]{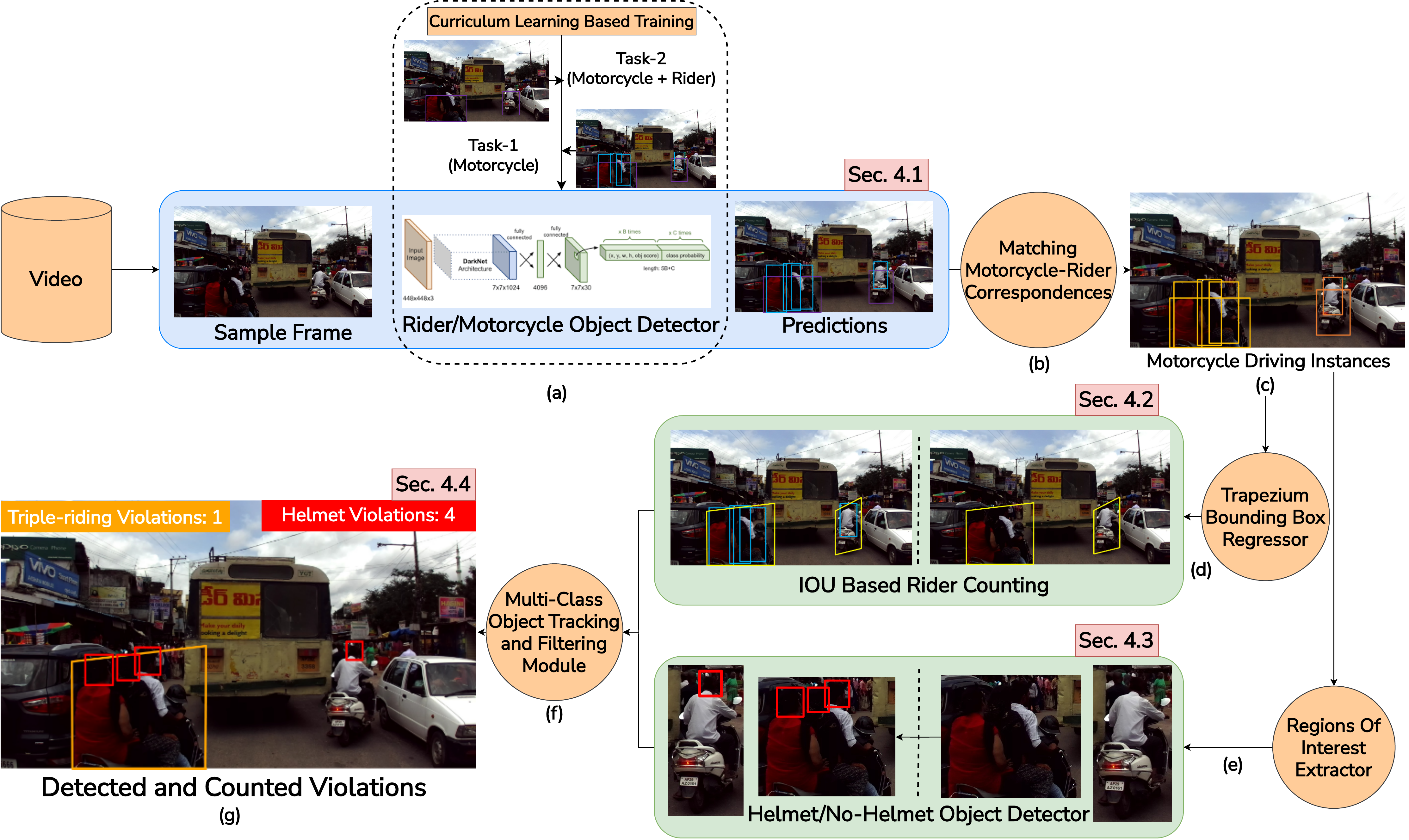} 
  \vspace{-7mm}
  \caption{Our pipeline: (a) Video frames are input to Curriculum Learning-based rider/motorcycle detector, (b) The intersecting motorcycles and riders are matched based on IOUs, (c) Each Rider-Motorcycle instance is shown in a different color, (d) Trapezium boxes from a regressor are passed to Triple-riding violation module, (e) Extracted ROIs are passed to a {\it helmet/no-helmet} detector, (f) Violations are tracked and counted over the video, (g) Violations/counts are shown in red and orange colors.}
  \label{Figure:3}\vspace{-2mm}
\end{figure*}
\section{The Pipeline}~\label{Section:Pipeline}
This section discusses the proposed pipeline, shown in ~\cref{Figure:3}, for identifying triple-riding and helmet violations. The videos from a camera mounted on a moving vehicle are input to a Curriculum Learning ({\sc cl}) based object detection module, predicting riders and motorcycles (refer to ~\cref{Figure:3} a). Thereafter, we send each motorcycle prediction and riders' predictions intersecting with it to the next stage (see ~\cref{Figure:3} b-c). A novel trapezium box regressor, shown in ~\cref{Figure:3} (d), then generates bounding boxes over each driving instance. Next, a triple-riding detection module calculates the number of riders within each trapezium box to identify the possibility of a triple-riding violation. A helmet violation detector also operates parallelly (see ~\cref{Figure:3} d), which works on the Regions Of Interest (ROIs) extracted from the previous detector's predictions (motorcycle and riders boxes). Finally, we modify the predictions from DeepSORT~\cite{wojke2017simple} to jointly track the rectangular {\it helmet/no-helmet} boxes and trapezium instance boxes as shown in ~\cref{Figure:3} (f-g). We now elaborate on each step of the proposed pipeline.
\subsection{Curriculum Learning-based Detector}~\label{Section:Pipeline(CL detector)}
A generic two-class {\it rider-motorcycle} detector performs poorly due to overlapping regions between a motorcycle and its riders. The overlapping regions lead to inter-class confusion and, in turn, lead to false negatives due to lower confidence of one or more overlapping predictions and Non-Maximum Suppression (NMS). As shown in ~\cref{Figure:3} (a), we propose a {\sc cl} based detector to avoid such issues. The model is first trained on only motorcycle instances and then retrained after adding rider instances. As we will see in  ~\cref{Section:Results}, the recall improves significantly compared to the generic model, as an outcome of a reduction in false positives. 
\begin{figure}[t]
  \centering
  \includegraphics[width=8.5cm]{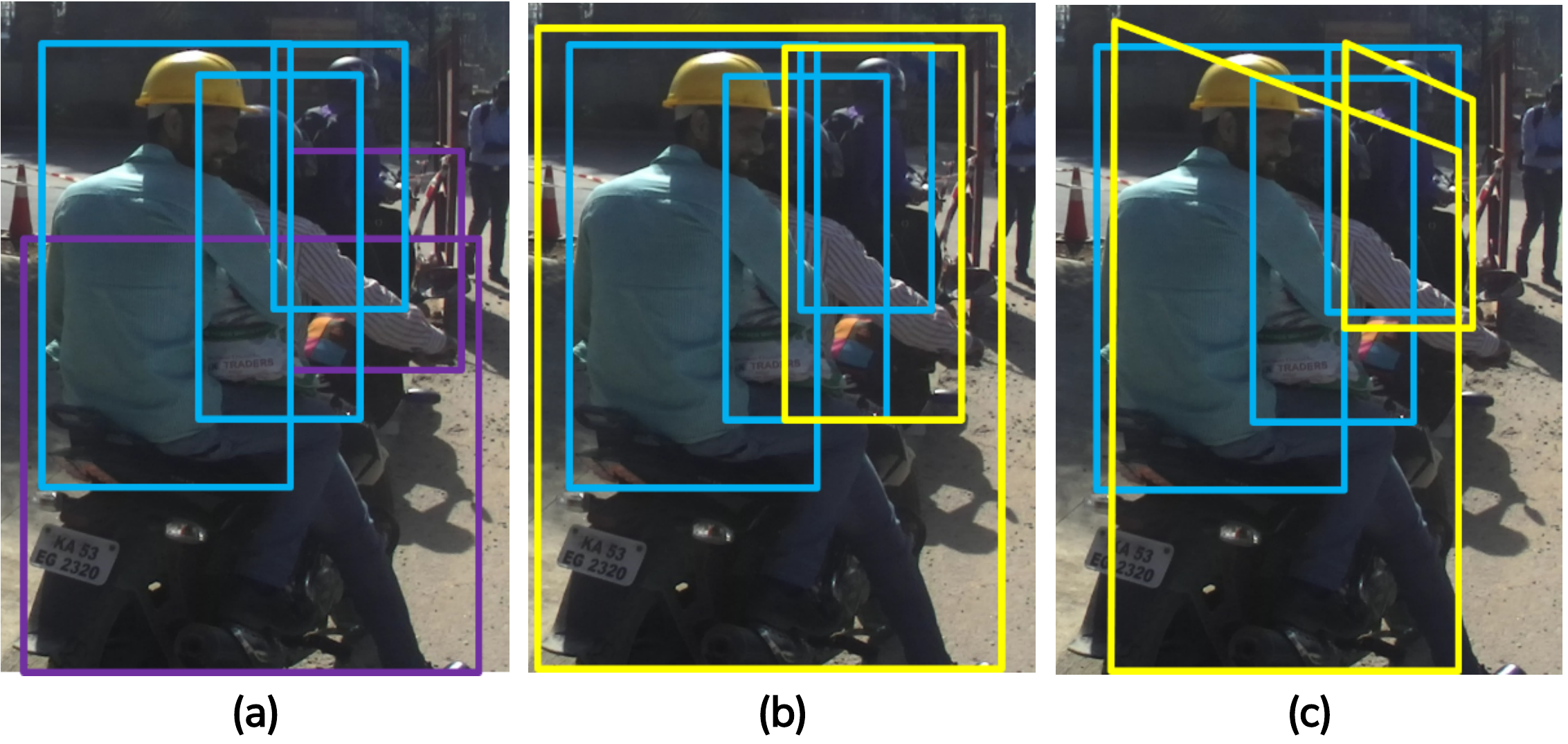}
  \vspace{-6mm}
  \caption{Different IOU-based correlation approaches to associate riders to their motorcycle. \textbf{Note:} (a) demonstrates the use of motorcycle boxes (purple) for IOU based association with rider boxes (blue), but the former have low IOU with all the latter, (b) demonstrates the use of instance boxes (yellow), but the box closer to the camera has high IOU with all the three riders, (c)~\textbf{[Proposed]} approach uses a trapezium-shaped instance box (yellow) has high IOU with it's two true riders and low IOU with the rider from another motorcycle.}
  \label{Figure:4}
\end{figure}
\begin{figure*}[t]
  \centering
  \vspace{-4 mm}
  \includegraphics[width=\linewidth]{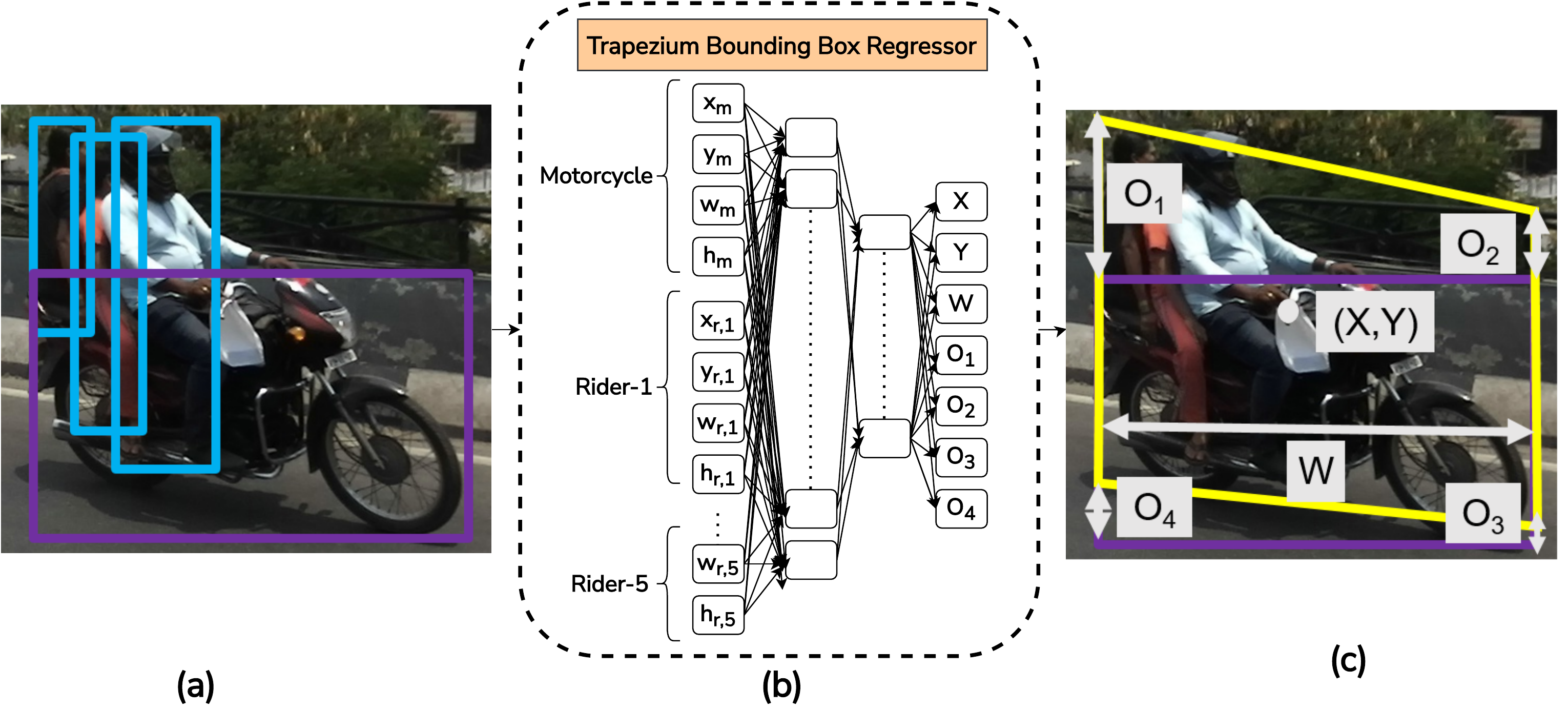}
  \vspace{-6.5mm}
  \caption{Input (a) and output (c) of the trapezium box regressor (b). Refer to  ~\cref{Section:Pipeline(TRV)} for additional details.}
  \label{Figure:5}\vspace{-2mm}
\end{figure*}

\subsection{Identifying Triple-riding Violations}~\label{Section:Pipeline(TRV)}
This section discusses the trapezium box regressor (see ~\cref{Figure:3} d) and the rider counting step for identifying triple-riding violations in a video frame. Each predicted motorcycle box and rider bounding boxes that intersect with it form the input to the trapezium regressor. The regressor detects driving instances for counting the rider boxes having the highest IOU with the obtained trapeziums; to identify the triple-riding violations. The resulting reduction in false-positive rider correspondences also leads to a massive $55.44\%$ precision gain for triple-riding violations in ~\cref{Table:2}.

\noindent
{\bf Trapezium Bounding Box Regressor:} 
Our data contains highly unstructured and crowded scenarios. Thus, as is illustrated in ~\cref{Figure:4} (a), relying on derived IOU-based correlations between riders and their motorcycles become erroneous for accurate identification of triple-riding violations. Similarly, drawing rectangular boxes over such correlated riders and motorcycles also becomes an issue, as in many cases, a significant portion of such boxes cover the background area and intersect with riders of other motorcycles, as shown in ~\cref{Figure:4} (b). Hence, we utilize novel trapezium bounding boxes which circumscribe the correlated riders and motorcycles more tightly than rectangles, as shown in ~\cref{Figure:4} (c). We train a two-layered deep network (see ~\cref{Figure:5} b) having $512$ and $256$ nodes in the respective hidden layers and use {\it tanh activation}  with {\it learning rate} of $0.001$. {\it Mean squared error }~\cite{allen1971mean} was used as loss function in regression. 

{\bf Input:} As shown in ~\cref{Figure:5} (a), corresponding motorcycle and riders' boxes are concatenated to form the input of the trapezium regressor. The number of inputs is fixed to $24$; the first four values correspond to the motorcycle box and the rest to its riders. If the number of riders is lower than $5$ (maximum number of rider space in input as also advocated by~\cite{lin2021positional}), then we initialize the remaining values to 0.

\textbf{Output:}
The center of gravity~\cite{nurnberg2013calculating} of the trapezium is taken as its central position (X, Y), the perpendicular distance between the parallel sides (vertical in our case) is taken as its width (W). For slopes of non-parallel sides, we take the offset of each corner of a trapezium from its corresponding motorcycle box corner as illustrated in ~\cref{Figure:5} (c).

\textbf{Equation of Trapezium's Center:}
\text Trapezium's centre, $(X,Y)$ which is defined by its $n~(=4)$ vertices $(x_0,y_0),(x_1,y_1),...(x_{n-1},y_{n-1})$ can be calculated by using the following equations. The vertices should be provided as input in clockwise or anti-clockwise order and the polygon should be closed such that the vertex $(x_0,y_0)$ is the same as the vertex $(x_n,y_n)$.:
\begin{equation}
    X = [\sum_{i=0}^{n-1}(x_{i} + x_{i+1})(x_{i}y_{i+1} - x_{i+1}y_{i})]/6A
\end{equation}
\begin{equation}
    Y = [\sum_{i=0}^{n-1}(y_{i} + y_{i+1})(x_{i}y_{i+1} - x_{i+1}y_{i})]/6A
\end{equation}
\text{Where A is a polygon's signed area:}
\begin{equation}
    A = [\sum_{i=0}^{n-1}(x_{i}y_{i+1} - x_{i+1}y_{i})]/2
    \label{eq:plygon_signed_area}
\end{equation}
\bigbreak

{\bf Counting Riders:} Counting the riders with boxes having the highest IOU with each trapezium bounding box are considered to be riding the motorcycle and therefore help identify violations.

One of the advantages of the trapezium-shaped boxes is that they are generic and can represent many other objects such as lane-markings and railway tracks recorded from a front view camera on a vehicle or train. It can benefit from the speed of detection methods and the accuracy of segmentation methods. We note that rider and motorcycle predictions obtained using semantic information are sufficient for the trapezium regressor. The trapeziums are robust and avoid bounding box area and region-based rules~\cite{mallela2021detection,saumya2020machine} for different conditions across images.
\begin{table*}[ht]
\small
\caption{Evaluation of Triple-riding Violation Detection (rider and motorcycle) \& Identification (counting riders on a bike). Note: {\sc cl} based YOLOv4 (v4) model performs better than the conventional two-class (non-{\sc cl}) v4 \& YOLOv3 (v3) models due to high IOU of riders with motorcycles. Trapezium-shaped boxes further boost the F-score.}\label{Table:1}\vspace{-2mm} \centering
\begin{tabular}{|c|c| c | c | c | c | c | c|}
  \hline
  \multirow{2}{*}{\textbf{S. No.}} & \multirow{2}{*}{\textbf{Method}} & \multicolumn{2}{|c|}{ \textbf{Rider-Motorcycle Detection}} & \textbf{Rider-Motorcycle Association} & \multicolumn{3}{|c|}{\textbf{Violation Identification Scores}} \\
  \cline{3-4}\cline{6-8}
  & & \textbf{Base} \textbf{Model} & \textbf{mAP} & \textbf{Approach} & \textbf{Precision} & \textbf{Recall} &  \textbf{F-score} \\
  \hline
  1 & Saumya et al.~\ \cite{saumya2020machine} & v3\_non-CL & 70.13\% & Euclidean Distance  &
  29.00\% & 50.00\% &  36.70.\%\\
  \hline
  2 & \multirow{4}{*}{Ours} & \multirow{4}{*}{v4\_non-CL}  & \multirow{4}{*}{73.62\%} & Euclidean Distance  &
  30.47\% & 51.26\% &  38.22\% \\
  3 & &  & & Rider-Motorcycle Box &
  33.73\% & 53.84\% &  41.47\% \\
  4 & & & &Rectangular-shaped Instance Box & 41.66\% & 57.69\% &  48.38\% \\
  5 & &  & &Trapezium-shaped Instance Box & 73.80\% & 59.61\% &  65.95\% \\
  \hline
  \textbf{6} & \textbf{Proposed} & \textbf{v4\_CL} & \textbf{82.61\%} &  \textbf{Trapezium-shaped Instance Box} & \textbf{84.44\%} & \textbf{73.07\%} &  \textbf{78.34\%} \\
  \hline
\end{tabular}
\vspace{-1mm}
\end{table*}
\begin{table*}
\small
\caption{Helmet Violation Detection and Identification Performance at Rider-level (rows 1-5) and Instance-level (rows 6-7). Note: {\sc cl} based YOLOv4 (v4) model trained on {\it helmet/no-helmet} classes performs poorer than the conventional two-class (non-{\sc cl}) v4 model due to low IOU between helmets and no-helmets. Refer ~\cref{Table:1} caption and detection performance for acronyms and {\it rider-motorcycle} mAP scores. N/A: Not Applicable.}\vspace{-2mm}\label{Table:2} \centering
\begin{tabular}{|c|c| c | c | c | c | c | c|}
  \hline
  \multirow{2}{*}{\textbf{S. No.}} & \multirow{2}{*}{\textbf{Method}} & \multicolumn{2}{|c|}{\textbf{Helmet/No-Helmet Detection}} & \textbf{ROI Extraction} & \multicolumn{3}{|c|}{\textbf{Violation Identification Scores}} \\
  \cline{3-4}\cline{6-8}
  & & \textbf{Base} \textbf{Model} & \textbf{mAP} & \textbf{Approach} & \textbf{Precision} & \textbf{Recall} &  \textbf{F-score} \\
  \hline
  1 & Rithish et al.~\cite{harish2021evaluate} & \multirow{3}{*}{v4\_non-CL} & \multirow{3}{*}{\textbf{90.00\%}}& Rider Instance Crop & 53.21\% & 81.02\% &  64.23\% \\
  2 & Ours & & & Upper Half Instance Crop & 49.9\% & 76.40\% &  60.36\%\\ 
  3 & Ours & & & Full Resolution Image & 71.36\% & 90.13\% &  79.65\% \\ \hline
  4 & Ours & v4\_CL & 83.5\% & Rider-Motorcycle Instance Crop & 68.56\% & 83.25\% &  75.19\% \\ \hline
  5 & \textbf{Proposed} & \textbf{v4\_non-CL} & \textbf{90.00\%} & \textbf{Rider-Motorcycle Instance Crop}  & \textbf{77.86\%} & \textbf{92.23\%} &  \textbf{84.43\%} \\
  \hline \hline
  6 & Chairat et al. \cite{Chairat_2020_WACV} & v3\_GoogleNet & N/A & Upper Half Instance Crop   & 84.61\% & 54.45\% &  66.25\%\\ \hline
  7 & \textbf{Proposed}  & \textbf{v4\_non-CL } & \textbf{90.00\%} & \textbf{Rider-Motorcycle Instance Crop}  & \textbf{99.01\%} & \textbf{95.23\%} &  \textbf{97.08\%} \\
  \hline
\end{tabular}
\end{table*}
\begin{figure*}[t]
  \centering
  \vspace{-2 mm}
  \includegraphics[width=0.8\linewidth]{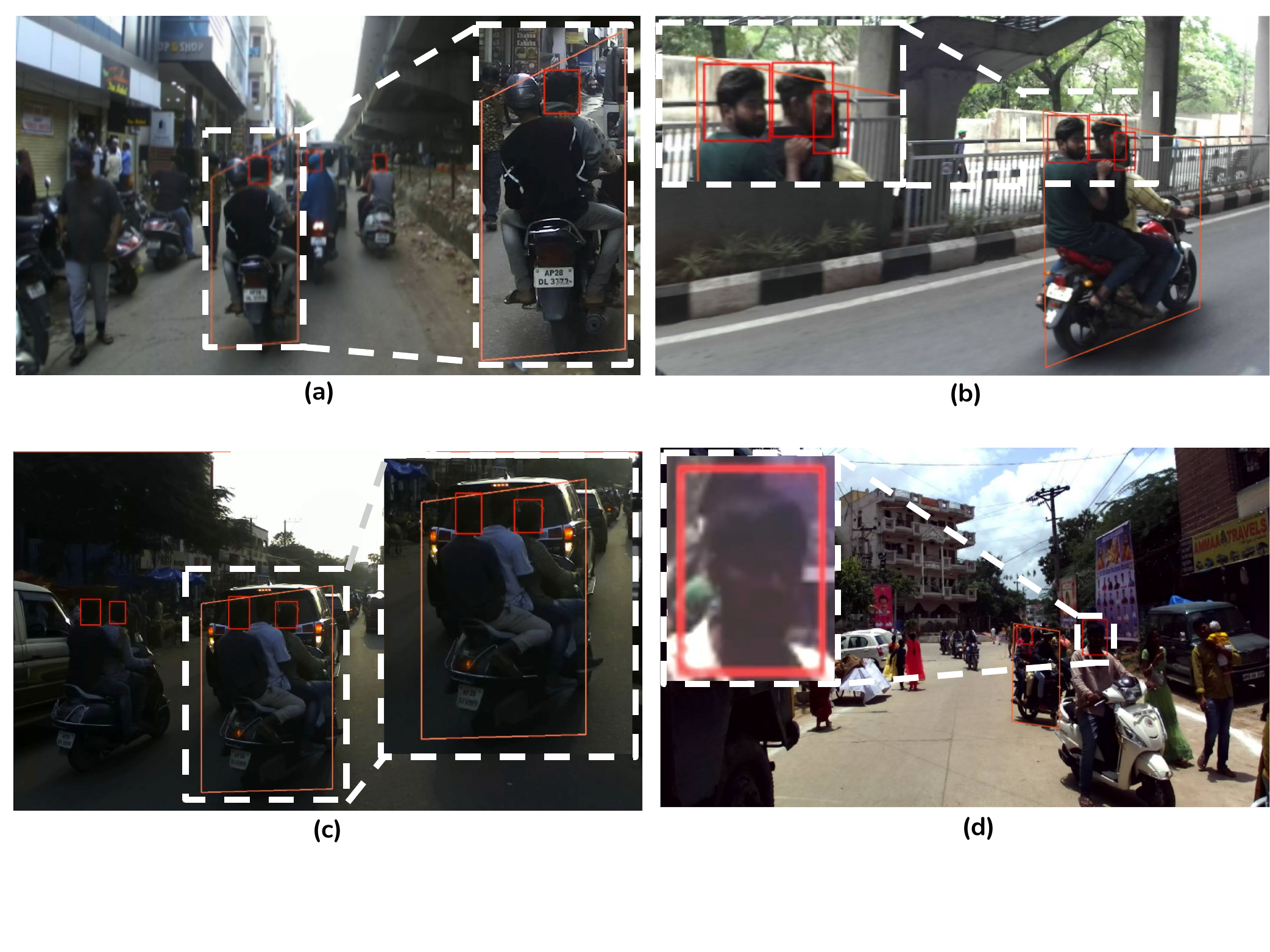}
  \vspace{-15mm}
  \caption{Qualitative results in complex scenarios: (a) Triple-riding violation with rider occlusions, (b) Helmet violations with head occlusions, (c) Triple-riding violation in a low light scene, (d) Helmet violation in a low light scene (top-left: crop enhanced for visualization).}
  \label{Figure:6}
\end{figure*}
\begin{figure*}[t]
  \centering
  \vspace{-2 mm}
  \includegraphics[width=0.8\linewidth]{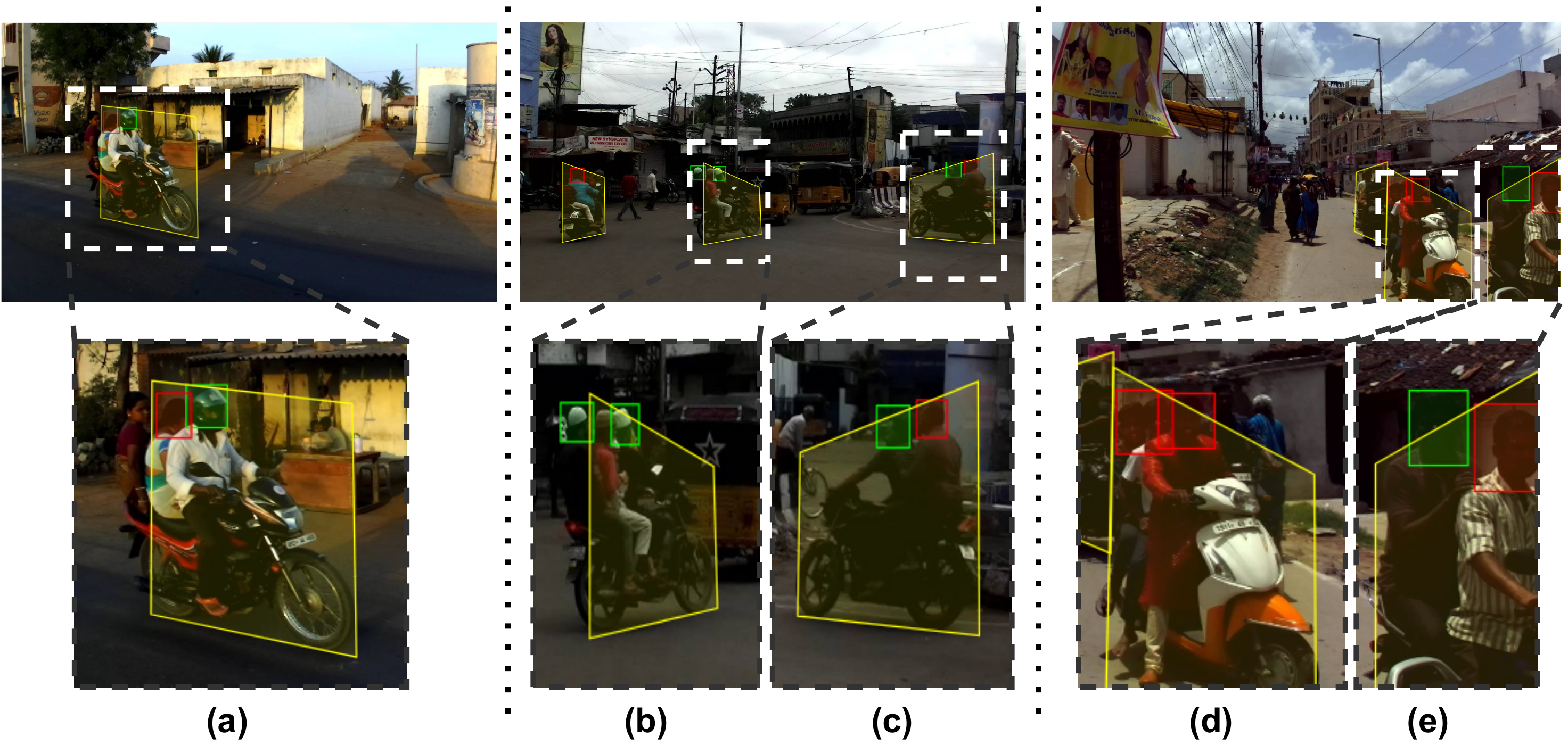}
  \vspace{-3mm}
  \caption{Failure cases. (a), (b), \& (d): riders with non-distinguishable backgrounds leading to missed triple-riding violations, (b), (c) \& (e): heads with caps and non-distinguishable backgrounds are sometimes falsely detected as helmets (green).}
  \label{Figure:9}
\end{figure*}
\subsection{Identifying Helmet Violations}~\label{Section:Pipeline(HV)}
As an input to the helmet violation model, the width of the predicted motorcycle and maximum height of predicted rider boxes of a particular motorcycle driving instance are used to extract the relevant ROIs. The cropped ROIs are extended by $10\%$ and then passed to the trained detector to predict the violations at the inference stage. Unlike motorcycle and rider classes, helmet and no-helmet classes have much lesser overlapping due to which a two-class YOLOv4 detector is trained (refer~\cref{Figure:3} e and~\cref{Figure:2} b).

For all detectors mentioned above, we use {\it binary cross entropy} and {\it CIoU (Complete Intersection over Union)} for classification and regression, respectively \cite{bochkovskiy2020yolov4}.

\subsection{Tracking}~\label{Section:Pipeline(Tracking)}
As shown in ~\cref{Figure:3} (f-g), we modify DeepSORT~\cite{wojke2017simple} predictions to jointly track {\it helmet/no-helmet} boxes and trapezium boxes in traffic videos. DeepSORT tracks only rectangular-shaped boxes, so the trapezium box is tracked via its corresponding motorcycle bounding box.
\section{Experiments and Results}~\label{Section:Results}
We use a dataset of $1281$ images containing $907$ riders without helmets and $98$ instances of triple-riding violations amongst $3313$ riders and $4573$ motorcycles. We use $70$$:$$30$ as a train:test split for our data. Our test set consists of $310$ images, where $260$ are from the filtered IDD dataset (refer to ~\cref{Section:Dataset}), and $50$ are randomly chosen images from web search to enrich the testing for rare triple-riding violations. It has $234$ riders with helmet violations and $42$ instances of triple-riding violations among $758$ riders and $685$ motorcycles. For the curriculum learning-based model, the initial learning rate is set to $0.001$ every time a class is added with a decay of 10. For the helmet violation detector as well, the learning rate is initialized with $0.001$ with the decay of $10$. We now present the results of our models on triple-riding and helmet violations, along with the previous works trained and tested on our data. We also present ablation studies, which show the performance boost by utilizing the proposed pre-processing and curriculum learning-based object detector.\\
\subsection{Comparison with Existing Methods}

{\bf Triple-riding Violations:} 
As shown in the first four columns of ~\cref{Table:1}, the mAP for the motorcycle and rider detections by the proposed curriculum learning ({\sc CL}) based YOLOv4 model
(see ~\cref{Section:Pipeline(CL detector)}) is over $8.9\%$ higher than the (non-{\sc cl}) YOLOv4 model as well as Saumya et al.~\cite{saumya2020machine}.

We now discuss the results on identifying triple-riding violations shown in the last four columns of ~\cref{Table:1}. The Euclidean distance-based rider-motorcycle association by Saumya et al.~\cite{saumya2020machine}, and the IOU-based approaches in ~\cref{Figure:4} (a) and (b) of ~\cref{Section:Pipeline(TRV)}, are not robust to false-positive riders from other motorcycles.  Hence, all these approaches have precision and F-Score lower than $50$ (rows 1-4 of ~\cref{Table:1}). However, using the proposed trapezium-shaped instance boxes as shown in ~\cref{Figure:4} (c) improves the F-score by $17.57\%$ as shown in ~\cref{Table:1} (row 5). Using the {\sc CL}-based base model on our proposed violation identification approach further improves the F-score by $12.39\%$, leading to the highest precision, recall, and F-score of $84.44\%$, $73.07\%$ and $78.34\%$. 

{\bf Helmet Violations:}
Prior works identify helmet violations in two ways;
i) Rider-level violations, which includes counting riders without a helmet, 
ii) Instance-level violations for motorcycles with at least one of their riders without a helmet. We outperform previous works at both levels.

Unlike the rider and motorcycle boxes, the helmet and no-helmet boxes have comparatively lower IOU with each other. Thus, using a {\sc CL} based detector to learn helmet and then no-helmet classes reduces the mAP from $90.00\%$ to $83.5\%$, as shown in columns 1-4, rows 1-4 of ~\cref{Table:2}. Therefore, as shown in rows 5 and 7, we propose to use the generic (non-{\sc cl}) YOLOv4 to detect helmet violations.

As mentioned in ~\cref{Section:Related Works}, there are many ways to provide the ROIs to the violation detection model. The background context present in the ROI has a pronounced effect on the detection performance. As shown in the last four columns of ~\cref{Table:2}, ROIs having complete {\it rider-motorcycle} instance (rows 5 and 7) provides enough background context when compared to rider-crops~\cite{harish2021evaluate}, upper body crops~\cite{Dahiya16,Shine2020,Singh2020,Chairat_2020_WACV,khan2020helmet}, or the full resolution images. We use ROIs obtained from predictions of the v4\_CL model in ~\cref{Table:1}. The proposed {\it rider-motorcycle} ROIs achieve the highest precision, recall, and F-score of $77.86\%$, $92.23\%$ and $84.43\%$. For instance-level violations, our approach significantly improves the F-score by $30.83\%$ compared to Chairat et al.~\cite{Chairat_2020_WACV} leading to precision, recall, and F-score of $99.01\%$, $95.23\%$ and $97.08\%$ as shown in the last two rows of ~\cref{Table:2}.

Qualitative results are shown in ~\cref{Figure:6},~\cref{Figure:9},~\cref{Figure:1} (c) and~\cref{Figure:3} (g).
~\cref{Figure:6} shows violations identified by our model in complex scenarios at different scales and orientations. The scenarios include triple-riding and helmet violations with (a) rider occlusions and (b) head occlusions. ~\cref{Figure:1} (c) and (d) also present qualitative results on both types of violations in the low light scenes. While analyzing the failure cases, we observe that our model sometimes fails in instances where 
(i) riders have non-distinguishable background as shown in~\cref{Figure:9} (a), (b) and (d), and (ii) heads with caps and non-distinguishable backgrounds, as shown in~\cref{Figure:9} (b) (c) and (e). 

{\bf Results on Demo Video:} A practical system should have high precision or low false positives, as it is not appropriate to call a non-violator a violator. Interestingly, our approach demonstrates impressive precision scores for the two violations in the demo video having four different traffic scenarios. The scenarios include low-light conditions, road junctions, conditions under the fly-over, and highway roads. Specifically, the proposed architecture achieves precision, recall, and F-score of i) $100.00\%$, $80.00\%$ and $88.88\%$ for triple rider violation and ii) $87.50\%$, $63.63\%$ and $73.67\%$ for helmet violation on the demo video. The demo video and the code is available on the github page.

\begin{figure}[t]
  \centering
  \vspace{-1 mm}
  \includegraphics[width=\linewidth]{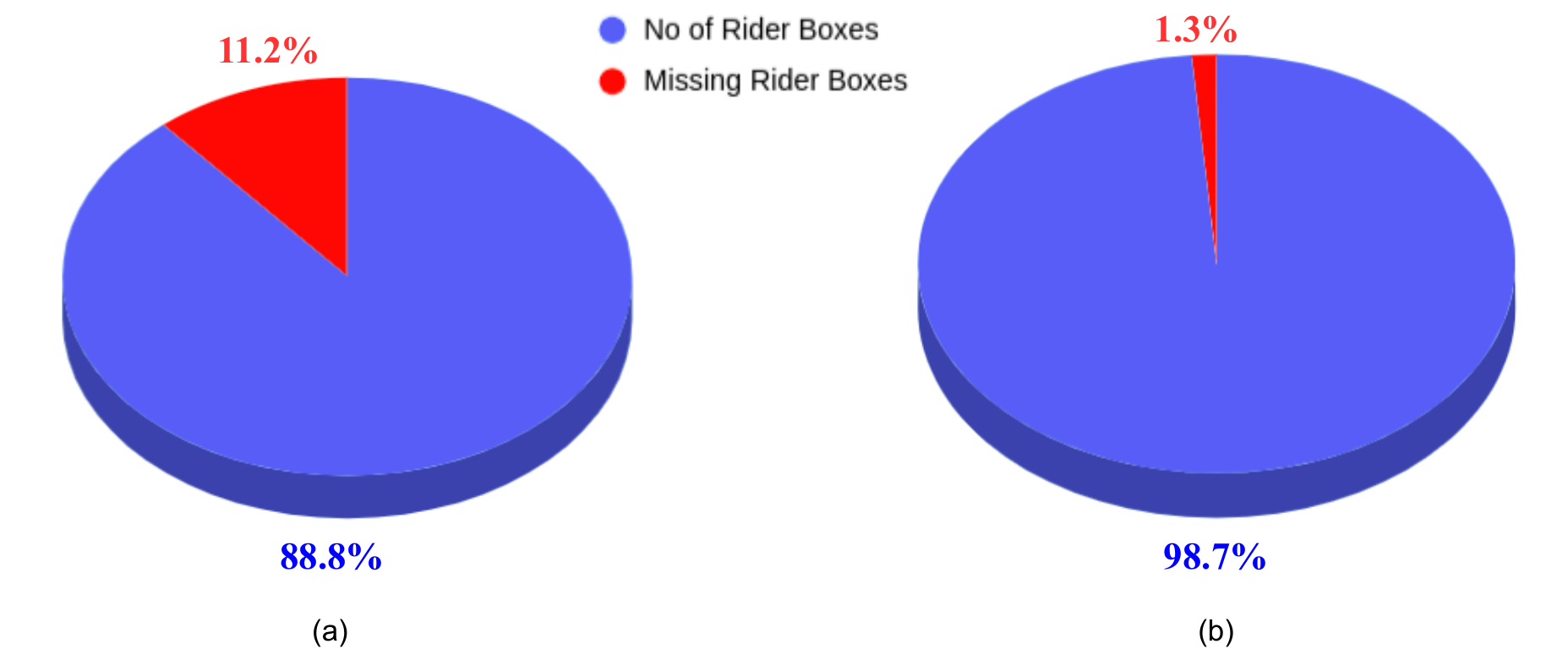}
  \vspace{-4.5mm}
  \caption{Effect of using Amodal Regressor (refer~\cref{Section:Dataset}): (a) and (b) show missing rider boxes before (11.2\%) and after (1.3\%) using the regressor, respectively.}
  \label{Figure:8}
\end{figure}
\begin{figure}[t]
  \centering
  \vspace{-2 mm}
  \includegraphics[width=\linewidth]{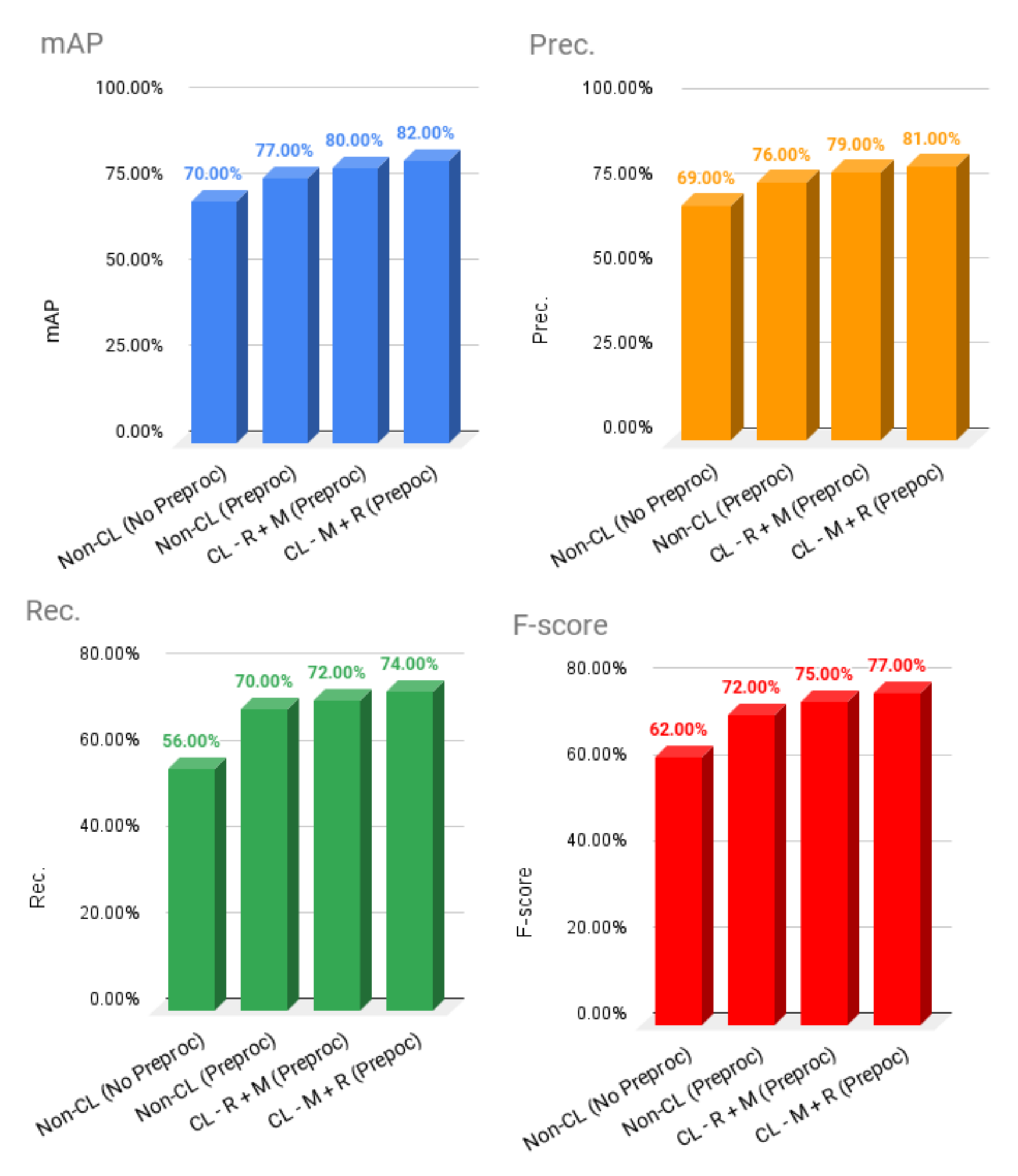}
  \vspace{-6.5mm}
  \caption{Detection Results: Effect of using proposed data pre-processing and curriculum learning.}
  \label{Figure:7}\vspace{-4mm}
\end{figure}
\subsection{Ablation Studies}

{\bf Data Pre-processing:}
As discussed in ~\cref{Section:Dataset}, we pre-process the IDD subset we use for training our model to include occlusions. More precisely, we introduced an amodal regressor that predicts a bounding box for an occluded rider using annotated {\it helmet/no-helmet} box (for the rider) as input. As shown in~\cref{Figure:8}, the percentage of missing rider boxes is significantly reduced by using the proposed methodology. It is important to note that i) the amodal regressor detects only one rider box for each helmet box, and ii) we keep the threshold for IOU between predicted and labeled rider boxes low and avoid all the false positives. We also analyze the effect of our pre-processing on the triple-riding violations task in~\cref{Figure:7}. We use a generic YOLOv4 model and trapezium boxes to associate the motorcycle with its riders for the analysis. Correcting the rider boxes has led to an increase in precision. False negatives or missing riders due to occlusion are also reduced, leading to improved recall. Overall, as shown in~\cref{Figure:7}, we obtain $7\%$ and $10\%$ gains in mAP and F-scores respectively due to the proposed data pre-processing techniques.

{\bf Trapezium Representation:} As seen in the previous section and~\cref{Table:1} (rows 4-5), the trapezium representation improved the F-Score of triple-riding violations by $17.57\%$ compared to the rectangle bounding box representation. The reduction in mis-association of riders to different motorcycles leads to an impressive $32.14\%$ improvement in precision. Trapeziums are a tighter fit of a motorcycle driving instance and hence cover less background and help in violation identification as explained in previous sections.

{\bf Curriculum Learning:} 
We utilize Curriculum Learning (CL) for training our object detector as discussed in~\cref{Section:Pipeline(CL detector)} and experiment with the class orders mentioned in the x-axis of~\cref{Figure:7}. We find that training first on the motorcycle and then on the rider class ({\bf M+R}) provides us the best results. A significant gain in recall compared to non-CL methods demonstrates the capability of the CL-based approach to robustly handle false negatives which arise due to overlapping predictions of the motorcycle and rider boxes.

\section{Conclusion}
~\label{Section:6}
This paper proposed a novel architecture capable of detecting, tracking, and counting triple-riding and helmet violations on crowded Asian streets using a camera mounted on a vehicle. Our approach jointly tackles both the violations. The proposed framework outperforms existing violation approaches because of its capability to handle rider occlusions, false negatives, and false-positive rides from other motorcycles in crowded scenarios. We demonstrate the effectiveness of using curriculum learning for motorcycle violations and trapezium bounding boxes instead of conventional rectangular boxes. Our work lays the foundation for utilizing such systems for increased road safety. It can be used for surveillance in any corner of the city without any considerable cost of a static camera network. In the future, we wish to explore the possibility of deploying our system to a distributed GPU setup for city-wide surveillance.

{\bf Acknowledgments:} We thank iHubData, the Technology Innovation Hub (TIH) at IIIT-Hyderabad for supporting this project.


\end{document}